\pdfoutput=1

\documentclass[11pt]{article}

\usepackage[final]{acl}

\usepackage{times}
\usepackage{latexsym}

\usepackage[T1]{fontenc}

\usepackage[utf8]{inputenc}

\usepackage{microtype}

\usepackage{inconsolata}

\usepackage{CJKutf8}
\usepackage{linguex}
\usepackage{graphicx}
\usepackage{amsmath}
\usepackage{multirow}
\usepackage{ulem}
\usepackage{booktabs}
\usepackage{diagbox}
\usepackage{pifont}

\definecolor{high}{rgb}{0.925,0.341,0.420} 
\definecolor{low}{rgb}{0.173,0.467,0.729}  
\definecolor{neutral}{rgb}{0.375,0.375,0.375}  
\newcommand{\high}[1]{\textcolor{high}{#1}}
\newcommand{\low}[1]{\textcolor{low}{#1}}
\newcommand{\neutral}[1]{\textcolor{neutral}{#1}}

\newcommand{\cmark}{\ding{51}}
\newcommand{\xmark}{\ding{55}}

\title{RecMind: Japanese Movie Recommendation Dialogue \\ with Seeker's Internal State}

\author
{
  Takashi Kodama$^1$, Hirokazu Kiyomaru$^1$, Yin Jou Huang$^1$, Sadao Kurohashi$^{1,2}$ \\
  $^1$Kyoto University, $^2$National Institute of Informatics \\
  \texttt{\{kodama, kiyomaru, huang, kuro\}@nlp.ist.i.kyoto-u.ac.jp}
}

\begin{document}
\maketitle
\begin{abstract}
  Humans pay careful attention to the interlocutor's internal state in dialogues.
For example, in recommendation dialogues, we make recommendations while estimating the seeker's internal state, such as his/her level of knowledge and interest. Since there are no existing annotated resources for the analysis, we constructed RecMind, a Japanese movie recommendation dialogue dataset with annotations of the seeker's internal state at the entity level. Each entity has a subjective label annotated by the seeker and an objective label annotated by the recommender. RecMind also features engaging dialogues with long seeker's utterances, enabling a detailed analysis of the seeker's internal state. Our analysis based on RecMind reveals that entities that the seeker has no knowledge about but has an interest in contribute to recommendation success. We also propose a response generation framework that explicitly considers the seeker's internal state, utilizing the chain-of-thought prompting. The human evaluation results show that our proposed method outperforms the baseline method in both consistency and the success of recommendations.
\end{abstract}

\section{Introduction}
In human dialogues, individuals pay careful attention to their interlocutor's internal state~\cite{chiba-etal-2014-user}, including their level of understanding and emotional state. Particularly in recommendation dialogues, where a recommender recommends something to a seeker, it is crucial to estimate what the seeker knows and what he/she is interested in. This understanding enables us to offer recommendations that better align with the seeker's preferences.

In the past few years, many large language models (LLMs) have been actively developed and have achieved remarkable performance in various natural language processing tasks~\cite{brown-etal-2020-gpt3,zhang2022opt,chowdhery2022palm,openai2023gpt4}. Current LLMs are able to generate human-like responses without specialized modules to consider the interlocutors. However, it remains an open question whether LLMs need to explicitly consider the seeker's internal state and how to effectively implement it. To answer this question, we need high-quality dialogue data with careful and fine-grained annotations of the seeker's internal state. Unfortunately, there are no existing recommendation dialogue datasets with internal state annotation.

One possible solution is to annotate existing recommendation dialogue datasets~\cite{li-etal-2018-redial,kang-etal-2019-recommendation,jia-etal-2022-e} with the seeker's internal state. However, the internal state labels annotated by a third party may not accurately reflect the actual state. \citet{kajiwara-etal-2021-wrime} point out a difference between the subjective emotions the document writer intends to convey and the objective emotions the document reader receives. Thus, it is necessary to collect new dialogues and have the seekers annotate them with the subjective internal state.

\begin{table*}[t!]
  \centering
  \footnotesize
  \begin{tabular}{p{0.5\textwidth}p{0.2\textwidth}rr}
  \toprule
  Dialogue & Entity & Knowledge & Interest \\ \midrule
  R1: What kind of movies do you usually watch? & - & - & - \\ \midrule
  S1: I often watch Japanese movies regardless of their genres. & Japanese movies & \high{\textit{H}} / \high{\textit{H}} & \high{\textit{H}} / \high{\textit{H}} \\ \midrule
  R2: OK, Japanese movies. Do you know an eccentric movie called ``DESTINY: The Tale of Kamakura''? & \multirow{2}{*}{Japanese movies} & \multirow{2}{*}{\high{\textit{H}} / \high{\textit{H}}} & \multirow{2}{*}{\high{\textit{H}} / \high{\textit{H}}} \\ \cmidrule{2-4}
   & eccentric movie &  \high{\textit{H}} / \low{\textit{L}} & \high{\textit{H}} / \neutral{\textit{N}} \\ \cmidrule{2-4}
   & ``DESTINY: The Tale of Kamakura'' & \neutral{\textit{N}} / \low{\textit{L}} & \high{\textit{H}} / \neutral{\textit{N}} \\ \midrule
  S2: I have never seen it. What kind of movie is it?  & - & - & - \\ \midrule
  R3: Through strange experiences around an unsuccessful mystery writer living in Kamakura and his new wife, they gradually discover their destiny. That's the story.
  It's interesting to see specters and ghosts. Because this is Kamakura, he said. & Strange experiences around an unsuccessful mystery writer living in Kamakura and his new wife & \multirow{5}{*}{\low{\textit{L}} / \low{\textit{L}}} & \multirow{5}{*}{\high{\textit{H}} / \neutral{\textit{N}}} \\ \cmidrule{2-4}
   & their destiny & \low{\textit{L}} / \low{\textit{L}} & \high{\textit{H}} / \low{\textit{L}} \\ \cmidrule{2-4}
   & specters and ghosts & \low{\textit{L}} / \low{\textit{L}} & \high{\textit{H}} / \low{\textit{L}} \\ \midrule
  S3: Sounds interesting. Could I see any street scenery of Kamakura? & street scenery of Kamakura & \high{\textit{H}} / \neutral{\textit{N}} & \high{\textit{H}} / \high{\textit{H}} \\ \midrule
  R4: The scenery of Kamakura does not appear so much. The main characters go to the land of Hades by train, so you can see a fantasy-like scene.
   & \multirow{2}{*}{The scenery of Kamakura} & \multirow{2}{*}{\high{\textit{H}} / \neutral{\textit{N}}} & \multirow{2}{*}{\high{\textit{H}} / \high{\textit{H}}} \\ \cmidrule{2-4}
   & the land of Hades & \neutral{\textit{N}} / \low{\textit{L}} & \high{\textit{H}} / \low{\textit{L}}\\ \cmidrule{2-4}
   & train & \neutral{\textit{N}} / \low{\textit{L}} & \high{\textit{H}} / \low{\textit{L}} \\ \cmidrule{2-4}
   & a fantasy-like scene & \high{\textit{H}} / \low{\textit{L}} & \high{\textit{H}} / \low{\textit{L}} \\ \midrule
  S4: I remembered that I had heard of this movie. Is Masato Sakai in it, I think? & - & - & - \\ \midrule
  R5: That's right. His new wife is played by Mitsuki Takahata. & His new wife & \high{\textit{H}} / \low{\textit{L}} & \high{\textit{H}} / \neutral{\textit{N}}\\ \cmidrule{2-4}
   & Mitsuki Takahata & \high{\textit{H}} / \high{\textit{H}} & \high{\textit{H}} / \neutral{\textit{N}}\\ \midrule
  S5: I see. Both are my favorite actors, so I would love to see them! & - & - & - \\
  \bottomrule
  \end{tabular}
  \caption{
    An example of RecMind. R and S denote the recommender and seeker, respectively.
    The Entity column lists the entities extracted from the dialogue.
    Each entity has subjective/objective labels for knowledge and interest.
    \high{\textit{H}}, \neutral{\textit{N}}, and \low{\textit{L}} denote \high{\textit{High}}, \neutral{\textit{Neutral}}, and \low{\textit{Low}}, respectively.
  }
  \label{tab:dialogue_example}
\end{table*}

To account for the aforementioned requirements, we constructed \textbf{RecMind}, a Japanese movie recommendation dialogue dataset. As illustrated in Table~\ref{tab:dialogue_example}\footnote{Examples of dialogues presented in this study are originally in Japanese and were translated by the authors.}, the recommender recommends movies based on the seeker's preferences in a multi-turn dialogue. We treat noun phrases as entities and annotate each of them with the seeker's level of knowledge and interest at three levels (\textit{High}, \textit{Neutral}, and \textit{Low}). In this process, the seeker assigns subjective labels to each entity, reflecting their own perception. Conversely, the recommender estimates the seeker's internal state based on the interactions with the seeker and assigns objective labels.

Our dataset also features high-quality dialogues with high dialogue enjoyment and recommendation success. Compared to JMRD~\cite{kodama-etal-2022-construction}, an existing Japanese movie recommendation dialogue dataset, our dataset stands out with longer seeker's utterances. This characteristic allows us to observe the internal states of a wide variety of entities.

Using the constructed dataset, we analyze the relationship between the seeker's internal state and the recommendation success. Our analysis reveals that entities without knowledge but with interest contribute to the successful recommendation. This finding suggests that the recommender should focus on topics or subjects that the seeker lacks knowledge of yet is interested in.

Based on the analysis results, we also propose a LLM-based response generation framework that explicitly considers the seeker's internal state. Specifically, we apply Chain-of-Thought prompting~\cite{wei2022chain} and estimate the seeker's internal state before generating a response. The human evaluation results demonstrate that our proposed method outperforms the baseline method, which does not explicitly consider the seeker's internal state, in both consistency and the success of recommendations.

In summary, our contributions are as follows.
\begin{itemize}
  \item We proposed RecMind, a Japanese movie recommendation dialogue dataset with subjective and objective annotations of the seeker's internal state at the entity level.
  \item We found entities the seeker has no knowledge about but has an interest in contributed to the successful recommendation.
  \item We proposed the response generation framework that explicitly considers the seeker's internal state, applying Chain-of-Thought prompting~\cite{wei2022chain}.
\end{itemize}

\section{Related Work}
Our work centers on the interlocutor's internal states in dialogues.
Besides, we describe the prior work on recommendation dialogue datasets.

\subsection{Internal State}
We focus on knowledge and interest as internal states in recommendation dialogues.
Here, we introduce previous studies that deal with knowledge and interest in dialogues.

\citet{miyazaki-etal-2013-estimating} proposed a method to estimate callers' levels of knowledge about particular themes (e.g., troubleshooting of products and services) in call center dialogues. Their annotations are performed at the dialogue level, whereas our dataset is annotated at the entity level. This allows for more fine-grained knowledge-state tracking and analysis. Inspired by the theory of mind~\cite{premack_woodruff_1978} and the common ground~\cite{clark1996using}, \citet{bara-etal-2021-mindcraft} created MINDCRAFT dataset which considers the user's knowledge for situated dialogue in collaborative tasks. Given the necessary knowledge and skills, two workers are asked to create a specific object together in the 3D virtual blocks world of Minecraft. Every predetermined time interval, players must answer a question about the common ground (e.g., ``Do you think the other player knows how to make YELLOW\_WOOL?''). In this study, we consider the user's knowledge in a more realistic dialogue that contains both chit-chat and recommendations.

Modeling interlocutor's interest has been actively studied in the field of recommendation dialogue~\cite{kang-etal-2019-recommendation,liu-etal-2020-towards-conversational,zhou-etal-2020-towards,jia-etal-2022-e}. In GoRecDialog~\cite{kang-etal-2019-recommendation}, each worker is given a set of five movies. The seeker's set represents their watching history, and the recommender's set represents the candidate movies to choose from. The recommender should recommend the appropriate movie among the candidates to the seeker. DuRecDial~\cite{liu-etal-2020-towards-conversational} is a recommendation dialogue dataset containing multiple dialogue types, such as question-answering and chit-chat. The recommender attempts to elicit the seeker's preferences, and the seeker responds based on a predefined user profile. These studies focus on the preferences for predefined objects (e.g., movies, user profiles). Our dataset differs in that we annotate all entities appearing in dialogues with the seeker's interest.

\subsection{Recommendation Dialogue Dataset}
Many previous studies have released recommendation dialogue datasets~\cite{li-etal-2018-redial,kang-etal-2019-recommendation,moon-etal-2019-opendialkg,liu-etal-2020-towards-conversational,zhou-etal-2020-towards,kodama-etal-2022-construction,jia-etal-2022-e}. INSPIRED~\cite{hayati-etal-2020-inspired} is a movie recommendation dialogue dataset associated with the recommendation strategy label (e.g., preference confirmation, personal experience). Using this annotation, they analyzed the recommender's strategies and pointed out that using sociable strategies (e.g., sharing personal opinions) more frequently leads to successful recommendations. We provide the labels of the seeker's internal state (i.e., knowledge and interest) from both the recommender's and the seeker's points of view.

\section{Data Collection} \label{sec:data_collection}
We collect data via crowdsourcing through a data supplier in Japan. In this section, we describe how we collect the RecMind dataset.

\subsection{Dialogue Collection Settings} \label{sec:dialogue_collection_setting}

\subsubsection{Workers} \label{sec:workers}
The two workers engaging in a dialogue have distinct roles: \textbf{recommender} and \textbf{seeker}. The recommenders recommend movies that align with the seeker's preferences, taking into account the seeker's current internal state. The seekers actively participate in the dialogue and ask questions if there is anything they do not know about what the recommender says.\footnote{For the detailed instructions distributed to the workers, see Appendix~\ref{sec:appendix_instruction}.}

It is assumed that recommendations from recommenders unfamiliar with movies might be short-sighted or less engaging because their knowledge about movies is sparse. Thus, we have two requirements for recommenders: (1) to be a movie enthusiast and (2) to watch an average of at least ten movies per year. We do not have any specific requirements for seekers.

\subsubsection{Tasks for Workers} \label{sec:tasks_for_workers}
Workers are required to complete the four specific tasks: dialogue, annotation of the seeker's internal state, annotation of external knowledge\footnote{In this study, knowledge means the seeker's internal state of knowledge, and external knowledge means the information the recommenders refer to in dialogues.}, and questionnaire.

\paragraph{Dialogue}
During a dialogue, the recommender suggests one or more movies to the seeker. Recommenders must actively gather enough information from the seeker through dialogue. They should also be attentive to the seeker's preferences rather than suggesting movies based on their own tastes. Meanwhile, seekers are encouraged to openly share their preferences and ask questions about any unknowns. Each participant is required to respond at least eight times.

\paragraph{Annotation of Seeker's Internal State}
The seekers annotate each entity in the dialogues with the subjective labels of the level of knowledge and interest, while the recommenders annotate the entity with the objective labels.

The options for knowledge are as follows:
\begin{description}
  \setlength{\itemsep}{-0.3em}
  \item[\textit{High}] The seeker has knowledge regarding the entity.
  \item[\textit{Neutral}] The entity cannot be said to be either \textit{High} or \textit{Low}. Or the level of knowledge for the entity cannot be judged from the given context.
  \item[\textit{Low}] The seeker does not have knowledge regarding the entity.
\end{description}

The options for interest are as follows:
\begin{description}
  \setlength{\itemsep}{-0.3em}
  \item[\textit{High}] The seeker is interested in the entity.
  \item[\textit{Neutral}] The entity cannot be said to be either \textit{High} or \textit{Low}. Or the level of interest for the entity cannot be judged from the given context.
  \item[\textit{Low}] The seeker is not interested in the entity.
\end{description}

In addition to the above three options, we introduce an additional option, denoted as \textit{Error}. This option is used when the extracted span is not a valid entity. We discard entities that the recommender or the seeker labeled as \textit{Error}. The annotation can be performed either during or after the dialogue.

\begin{table*}[t!]
  \centering
  \footnotesize
  \begin{tabular}{lll}
  \toprule
  & Question & Choice \\ \midrule
  Q1 & How many movies do you watch per year? & 5: 20 or more, 4: 10 to 19, 3: 5 to 9, 2: 3 to 4, 1: 2 or less \\ \midrule
  \multirow{5}{*}{Q2} & \multirow{5}{55mm}{Do you know the movie you recommended? (for recommenders)\\ Do you know the movie that was recommended? (for seekers)} & 5: have watched the movie and remembered the contents well \\
  & & 4: have watched the movie and remembered some of the contents \\
  & & 3: have never watched the movie but know the plots \\
  & & 2: have never watched the movie and know only the title \\
  & & 1: do not know at all \\ \midrule
  \multirow{2}{*}{Q3} & \multirow{2}{*}{Did you enjoy the dialogue?} & 5: agree, 4: somewhat agree, 3: neutral, \\
  & & 2: somewhat disagree, 1: disagree \\ \midrule
  \multirow{4}{*}{Q4} & \multirow{4}{55mm}{Do you think you have recommended the movie well? (for recommenders) \\ Do you want to watch the recommended movie? (for seekers)} & \\
  & & 5: agree, 4: somewhat agree, 3: neutral, \\
  & & 2: somewhat disagree, 1: disagree \\
  & & \\
  \bottomrule
  \end{tabular}
  \caption{Questions and choices of the questionnaire. The number at the beginning of each choice indicates the score for that choice.}
  \label{tab:questionnaire}
\end{table*}

\paragraph{Annotation of External Knowledge}
Following the previous research on knowledge-grounded dialogues~\cite{dinan-etal-2018-wizard,wu-etal-2019-proactive}, recommenders annotate their own utterances with the piece of external knowledge when they refer to it. The annotation is not required for utterances that do not refer to external knowledge, such as greetings and utterances containing the personal knowledge of the recommenders.
However, the recommenders are required to always annotate their utterances with the title of the recommended movies when mentioning them.\footnote{For dialogues missing the annotation of the recommended movies, authors read the dialogues and annotated them with the movie titles.}
This is to track recommended movies in the dialogues.

\paragraph{Questionnaire}
After the dialogue, workers answer the questionnaire shown in Table~\ref{tab:questionnaire}. We assign a score of 5 to 1 to each choice for each question.

\subsection{Dialogue Collection System}
We develop a web-based dialogue collection system for data collection.\footnote{Figures~\ref{fig:system_rec} and \ref{fig:system_sek} show the screenshots of the recommender's and the seeker's chatrooms, respectively.} This system is an extension of ChatCollectionFramework\footnote{\url{https://github.com/ku-nlp/ChatCollectionFramework}}, by adding a movie search tool and an internal state annotation tool.

\subsubsection{Movie Search Tool}
We create a movie search tool to assist recommenders in dialogues.
We first curate 2,317 popular movie titles and their genres from a Japanese movie information website, Yahoo! Movies.\footnote{\url{https://movies.yahoo.co.jp/}}
We then collect metadata for each movie from Wikipedia.
Metadata consists of the title, release date, running time, directors, casts, original work, theme song, production country, box office, and plots.\footnote{Some metadata may be missing.}
Additionally, we include reviews for 261 movies from JMRD~\cite{kodama-etal-2022-construction} as part of the metadata.

During dialogue collection, recommenders use this tool to search and check movie information. Searching can be done by genres or text-based queries. We save the search log with the corresponding recommender's utterance as one of the records of the recommender's behaviors. When sending an utterance, recommenders can annotate it with the referred external knowledge by clicking the checkbox on the side of each piece of external knowledge. This tool is only displayed on the recommender's screen, that is, the seekers cannot see movie information.

\subsubsection{Internal State Annotation Tool}
The internal state annotation tool displays the entities to be annotated on the screen of both the recommenders and the seekers. Entities are automatically extracted from utterances to reduce the load of workers. We consider noun phrases as entities. Modifiers are extracted together to make it easier to grasp their meanings. We use linguistic features from the Japanese morphological analyzer Juman++~\cite{morita-etal-2015-morphological,tolmachev-etal-2018-juman} and the Japanese syntactic analyzer KNP~\cite{kurohashi-nagao-1994-syntactic} for entity extraction.

\begin{table}[t]
  \centering
  \small
  \begin{tabular}{ll}
  \toprule
    \# dialogues & 1,201 \\
    \# utterances (R / S) & 10,697 / 10,317 \\
    Avg. \# utterances per dialogue & 17.5 \\
    \# movies & 739 \\
    \# workers (R / S) & 27 / 46 \\
    \# searches & 5,596 \\
    \# external knowledge & 5,250 \\
    \# entities (knowledge / interest) & 52,586 / 52,246 \\
  \bottomrule
  \end{tabular}
  \caption{Statistics of RecMind. R and S denote recommender and seeker, respectively.}
  \label{tab:statistics}
\end{table}

\begin{table*}[t]
  \centering
  \small
  \begin{tabular}{l|cccccccc|ccr}
  \toprule
  & \multicolumn{2}{c}{Q1} & \multicolumn{2}{c}{Q2} & \multicolumn{2}{c}{Q3 ($\uparrow$)} & \multicolumn{2}{c}{Q4 ($\uparrow$)} & \multicolumn{2}{|c}{Words ($\uparrow$)} & Ext. K. ($\downarrow$) \\ \midrule
  & R & S & R & S & R & S & R & S & R & S & - \\ \midrule
  JMRD & - & - & 3.94 & 2.72 & 4.00 & 3.83 & 4.01 & 3.82 & 23.80 & \hfill6.87 & 1.24 \\
  RecMind (non-enthusiasts) & 2.57 & 3.66 & 3.80 & 1.58 & 3.99 & 4.27 & 3.61 & 4.47 & \textbf{41.90} & \textbf{31.48} & 0.75 \\ \midrule
  RecMind & 4.73 & 3.54 & 3.17 & 1.79 & \textbf{4.29} & \textbf{4.42} & \textbf{4.27} & \textbf{4.51} & 41.07 & 31.08 & \textbf{0.49} \\
  \bottomrule
  \end{tabular}
  \caption{Results of the questionnaire and the comparison with JMRD. ``Words'' indicates the average number of words per utterance and ``Ext. K.'' indicates the average use count of external knowledge per recommender's utterance. R and S denote recommender and seeker, respectively. ``non-enthusiasts'' means the results of the dialogue collection by the recommenders who are not movie enthusiasts. Best results are in bold. The scores for Q1 and Q2 are not bolded because a higher (or lower) score does not imply superiority of any kind.}
  \label{tab:comparison_with_jmrd}
\end{table*}

\subsection{Statistics}
\subsubsection{Dialogue and Questionnaire}

Table~\ref{tab:statistics} shows the statistics of RecMind.
We collected 1,201 dialogues consisting of an average of 17.5 utterances.
739 movies were used in the dialogues, indicating our dataset contains diverse recommendation dialogues.

The bottom row in Table~\ref{tab:comparison_with_jmrd} shows the questionnaire results. The results of Q2 show that the recommenders often recommend movies that the seeker does not know.

\paragraph{Comparison with JMRD}
Table~\ref{tab:comparison_with_jmrd} also shows the comparison results with JMRD~\cite{kodama-etal-2022-construction}, a knowledge-grounded recommendation dialogue in the same language and domain. The result of Q3 shows that the recommendation process is more enjoyable for both recommenders and seekers in our dataset. Regarding Q4, the result shows that our recommendations are more successful. Notably, the average score of Q4 by seekers improved from 3.82 to 4.51, highlighting that our dialogues are high-quality recommendation dialogue.

We next compare the average number of words per utterance. The results demonstrate that our dataset has longer utterances than JMRD. Especially, the seeker's utterances of RecMind are more than four times longer than those of JMRD, which could facilitate the analysis of the seeker's internal state. We additionally compare the average count of external knowledge use per recommender's utterance and observe a decrease from 1.24 to 0.75 in our dataset. This decrease is because we did not mandate recommenders to use external knowledge, except when mentioning movie titles. We believe that it is unnecessary to link external knowledge to every utterance because humans only refer to external knowledge when necessary.

\paragraph{Influence of Recommender's Movie Knowledge}
As noted in Section~\ref{sec:workers}, we recruited movie enthusiasts who watched ten movies or more per year as recommenders. To verify the effectiveness of this recruitment, we collected 74 dialogues from recommenders who watched fewer than ten movies per year. This data collection followed the same methodology as described in Section~\ref{sec:dialogue_collection_setting}, except for the number of movies the recommenders watched.

Table \ref{tab:comparison_with_jmrd} shows the results. The average score of Q3 by seekers decreased from 4.42 to 4.27, and that of Q4 from 4.51 to 4.47. Furthermore, the scores for Q3 and Q4 by recommenders, which means self-evaluation, also decreased from 4.29 to 3.99 and from 4.27 to 3.61, respectively. These results indicate that movie enthusiasts are likely to deliver more enjoyable dialogues and recommend successfully.

While the length of utterances is comparable, the number of external knowledge used increases from 0.49 to 0.75. This is because the recommenders who are not movie enthusiasts tend to rely on external knowledge more frequently to compensate for their lack of knowledge about movies.

\subsubsection{Internal State}  \label{sec:internal_state}
\begin{table}[t]
  \centering
  \small
  \begin{tabular}{l|rrr|r}
  \toprule
  \diagbox[width=1.5cm]{Obj.}{Sub.} & \textit{High} & \textit{Neutral} & \textit{Low} & Total \\ \midrule
  \textit{High} & 20,664 & 3,084 & 4,794 & 28,542 \\
  \textit{Neutral} & 6,737 & 1,791 & 3,583 & 12,111 \\
  \textit{Low} & 5,154 & 1,502 & 5,277 & 11,933 \\ \midrule
  Total & 32,555 & 6,377 & 13,654 & - \\ \bottomrule
  \end{tabular}
  \caption{Statistics of knowledge annotation.}
  \label{tab:stats_knowledge}
\end{table}

\begin{table}[t]
  \centering
  \small
  \begin{tabular}{l|rrr|r}
  \toprule
  \diagbox[width=1.5cm]{Obj.}{Sub.} & \textit{High} & \textit{Neutral} & \textit{Low} & Total \\ \midrule
  \textit{High} & 28,244 & 4,338 & 746 & 33,328 \\
  \textit{Neutral} & 11,838 & 3,716 & 1,018 & 16,572 \\
  \textit{Low} & 1,346 & 549 & 451 & 2,346 \\ \midrule
  Total & 41,428 & 8,603 & 2,215 & - \\ \bottomrule
  \end{tabular}
  \caption{Statistics of interest annotation.}
  \label{tab:stats_interest}
\end{table}

RecMind has 52,586 and 52,246 entities annotated with the seeker's knowledge and interest, respectively. Tables~\ref{tab:stats_knowledge} and \ref{tab:stats_interest} show the statistics of the seeker's internal state annotations. For subjective knowledge labels, \textit{High} is the most common, followed by \textit{Low}. The distribution for subjective interest labels is more imbalanced than knowledge labels with \textit{High} being particularly dominant. This is probably because recommenders usually advance a dialogue toward topics of interest to the seekers. For objective labels, the number of \textit{Neutral} labels increases in both knowledge and interest. This is because it is difficult for recommenders to judge the seeker's internal state of some entities.

We calculate the agreement and Pearson correlation between the subjective and objective labels. The agreement is 0.53 for knowledge and 0.62 for interest labels, and the Pearson correlation is 0.27 for knowledge and 0.21 for interest. This result suggests that it is difficult to substitute subjective labels with objective ones.

\paragraph{Relationship between Knowledge and Interest}
We explore the correlation between subjective knowledge and interest labels for the same entities. The Pearson correlation coefficient is 0.12, indicating no correlation. This result means that knowledge and interest represent different facets of the internal state.

\begin{figure*}[t]
  \centering
  \includegraphics[width=\textwidth]{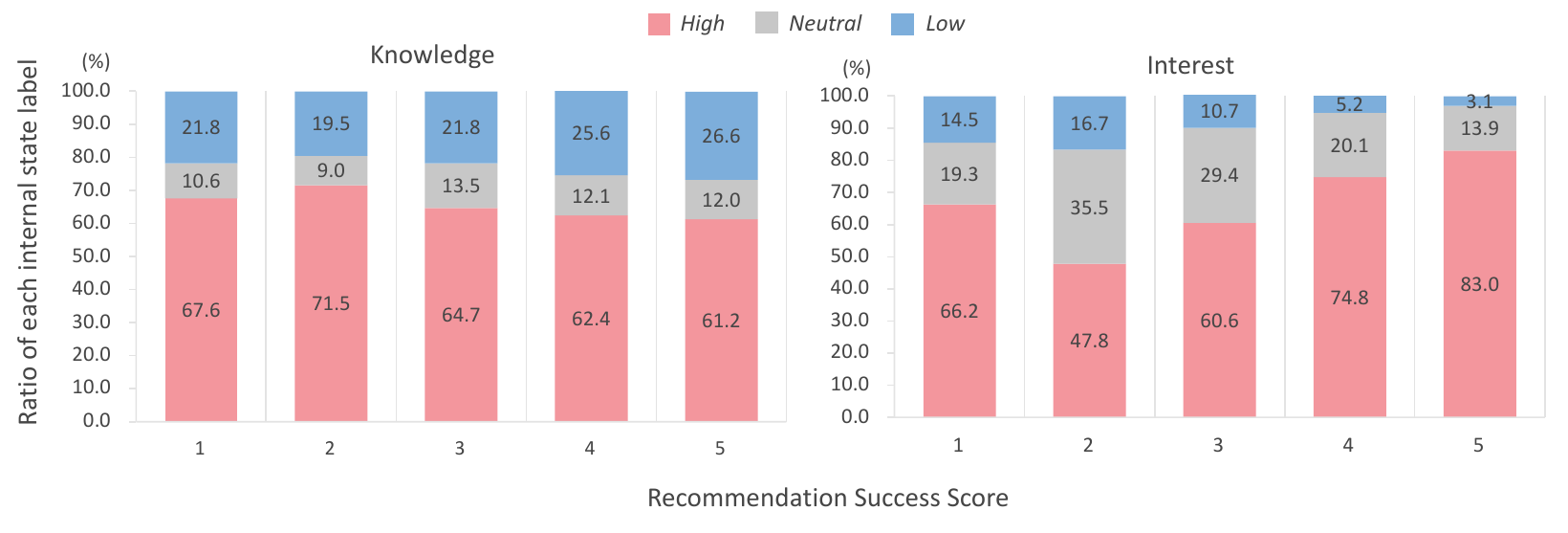}
  \caption{Relationship between recommendation success score and the ratio of each internal state label.}
  \label{fig:success}
\end{figure*}

\paragraph{Contribution of Seeker's Internal State to Recommendation Success}
We investigate the relationship between the subjective seeker's internal state and recommendation success at the dialogue level. We use the seeker's answer to Q4 (i.e., ``Do you want to watch the recommended movie?'') as an indication of recommendation success. Figure~\ref{fig:success} shows that dialogues with high recommendation success scores tend to have more \textit{Low} knowledge entities. For interest, on the other hand, dialogues with high recommendation success scores tend to have more \textit{High} interest entities.

We next analyze the dialogues with entities of \textit{Low} knowledge and \textit{High} interest in comparison with those dialogues without these kinds of entities. The average recommendation success score for the former dialogues is 4.59, while that for the latter dialogues is 4.18. Student's \textit{t}-test result reveals that the difference is statistically significant at the $p = 0.05$ level. The above analysis results indicate it is important in recommendation dialogues to identify and mention the topics where the seeker has no knowledge but has an interest.

\begin{table}[t]
  \centering
  \small
  \begin{tabular}{ll|rr}
  \toprule
  Knowledge & Interest & \cmark & \xmark \\ \midrule
  \textit{High} & \textit{High} & \textbf{3.61}\hspace{1mm} & \textbf{3.61} \\
  \textit{High} & \textit{Low} & 3.59\hspace{1mm} & \textbf{3.61} \\
  \textit{Low} & \textit{High} & \textbf{3.72}* & 3.53 \\
  \textit{Low} & \textit{Low} & 3.56\hspace{1mm} & \textbf{3.61} \\
  \bottomrule
  \end{tabular}
  \caption{Difference in recommendation success by each entity. \cmark and \xmark\hspace{0.5mm} denote the presence and absence of the entity in the utterance, respectively. The asterisk (*) indicates that the difference is statistically significant at the $p = 0.05$ level. Wilcoxon rank-sum test is used as a statistical test.}
  \label{tab:utterance_level_analysis}
\end{table}

Next, we explore the relationship between the subjective seeker's internal state and recommendation success at the utterance level for detailed analysis. To this end, we randomly selected 1,000 pairs of recommender's utterances and preceding dialogue context from our constructed dataset. We then ask crowdworkers to evaluate whether the utterance makes the interlocutor interested in watching a movie, using a 5-point Likert scale (5 is the best). Three workers evaluate each utterance, and the scores are averaged. Table~\ref{tab:utterance_level_analysis} shows the results. The score is high when the recommender's utterance includes entities with \textit{Low} knowledge and \textit{High} interest. The above results confirm that the recommender can effectively recommend by mentioning entities the seeker does not know but is interested in, even at the utterance level.

\section{Experiment}

\begin{figure*}[t!]
  \centering
  \includegraphics[width=\textwidth]{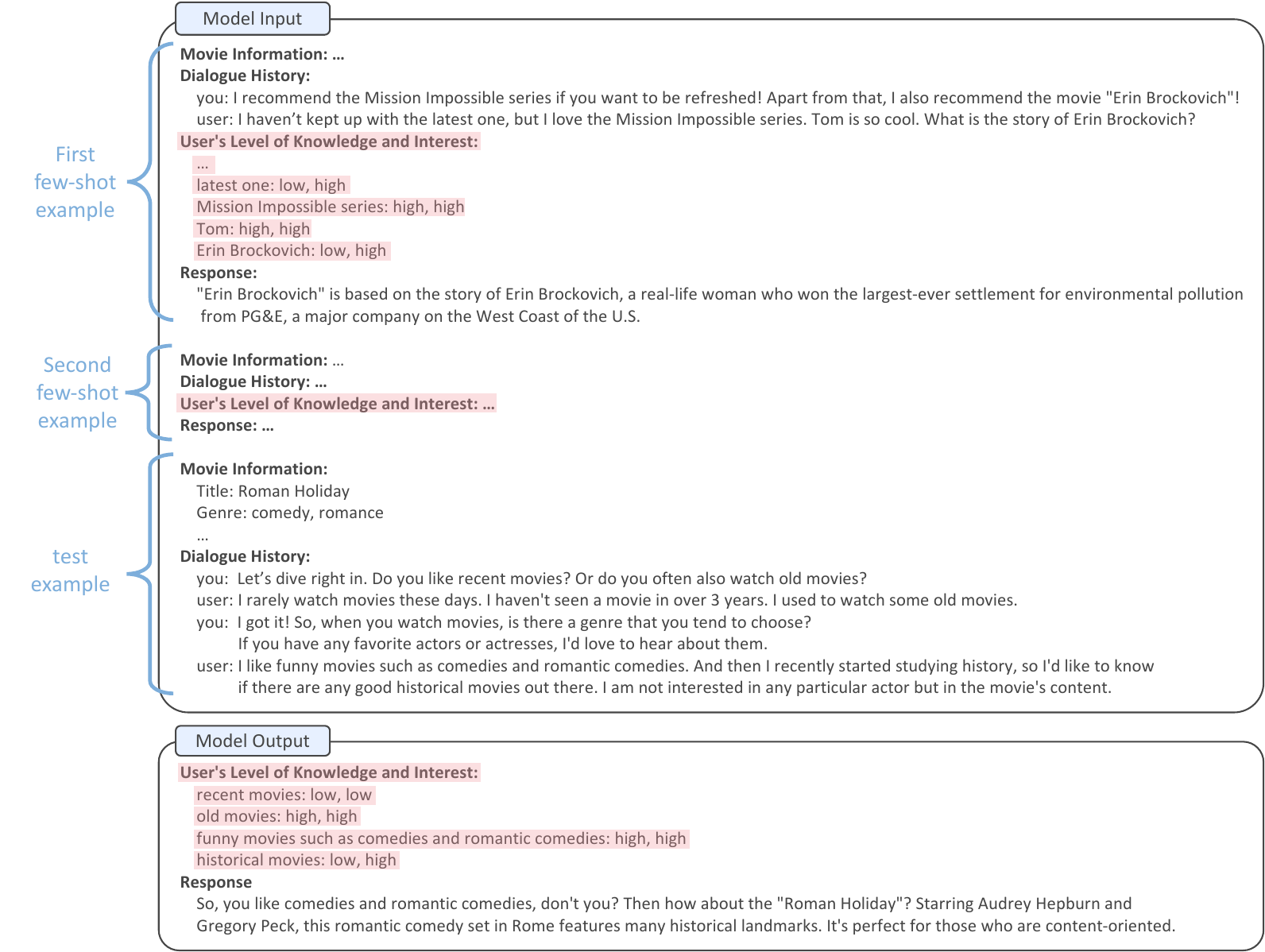}
  \caption{Overview of our proposed method. The internal state estimation, which is highlighted, is performed only for the proposed method and not for the baseline method.}
  \label{fig:proposed}
\end{figure*}

The analysis in Section~\ref{sec:internal_state} suggests the importance of understanding the seeker's internal state at the entity level. Thus, we propose a response generation framework that explicitly considers the seeker's internal state at the entity level. In this section, we describe our proposed method and verify its effectiveness.

\subsection{Proposed Method}
We propose a LLM-based response generation framework that explicitly considers the seeker's internal state labels.
Specifically, we apply Chain-of-Thought prompting~\cite{wei2022chain} to our task and estimate the seeker's internal state prior to generating a response.

Figure~\ref{fig:proposed} shows an overview of our proposed method.
Given the movie information and the dialogue history as the input for the model, the model first extracts target entities. It then estimates the seeker's level of knowledge and interest in each entity at three levels: \textit{High}, \textit{Neutral}, and \textit{Low}, respectively. Finally, the model generates a response referring to the estimated internal state.

\subsection{Experimental Settings}

\subsubsection{Model} \label{sec:model}

We use GPT-4 (\texttt{gpt-4-0613})~\cite{openai2023gpt4}, which achieves outstanding performance on various language-related tasks, as the base model. We selected GPT-4 because of its remarkable performance in JGLUE~\cite{kurihara-etal-2022-jglue}, the general natural language understanding benchmark for Japanese.\footnote{\href{http://nejumi.ai/}{http://nejumi.ai/}}

We compare the following four methods with a baseline that generates responses without estimating the seeker's internal state:
\begin{description}
  \item[CoT (sub): ] uses the subjective seeker's internal state labels in the few-shot examples.
  \item[CoT (obj): ] uses the objective seeker's internal state labels in the few-shot examples.
  \item[CoT (sub, gold): ] uses CoT (sub) method but is given the correct labels of the seeker's internal state in the test example, and only the response generation is performed. This method is for the purposes of an ablation study.
  \item[CoT (obj, gold): ] uses CoT (obj) method but is given the correct labels of the seeker's internal state in the test example, and only the response generation is performed. This method is for the purposes of an ablation study.
\end{description}

\subsubsection{Dataset}
We randomly split the collected dialogues into 85\%:15\% for training and test data, respectively. We selected the candidates for few-shot examples from the training data based on the following two criteria: (1) including all types of entity labels for knowledge and interest within the dialogue context, and (2) ensuring that the response incorporates an entity with \textit{Low} knowledge and \textit{High} interest. The second constraint is based on the findings in Section~\ref{sec:internal_state}, and was established to use higher-quality responses as few-shot examples. Consequently, we obtained 217 few-shot examples for CoT (sub) and 150 few-shot examples for CoT (obj). As for the test example, we randomly selected 500 examples from the test split only using the first criterion. For each test example, we then randomly chose two few-shot examples from the candidate pool.

\begin{table*}[t]
  \centering
  \small
  \begin{tabular}{l|rrrrr}
  \toprule
  \multirow{2}{*}{Model} & \multirow{2}{*}{Consistency} & Seeker's & Seeker's & Tailored & Recommendation \\
  & & Knowledge & Interest & Information & Success \\ \midrule
  CoT (sub) & 52.2* & 51.5\hspace{1.2mm} & 52.5* & 51.4\hspace{1.2mm} & 52.1* \\
  CoT (obj) & 51.4\hspace{1.2mm} & 52.1* & 52.2* & 52.3* & 51.3\hspace{1.2mm} \\ \midrule
  CoT (sub, gold) & 54.5* & 54.2* & 54.8* & 55.0* & 56.0* \\
  CoT (obj, gold) & 53.0* & 51.6\hspace{1.2mm} & 53.0* & 52.7* & 53.5* \\
  \bottomrule
  \end{tabular}
  \caption{Results of the response generation. The asterisk (*) indicates that the difference is statistically significant at the $p = 0.05$ level using a binomial test.}
  \label{tab:human_evaluation_result}
\end{table*}

\subsection{Result}
In this study, we conduct a human evaluation to assess the quality of the responses generated by the proposed methods. Specifically, we present the responses of each method in Section~\ref{sec:model} and the baseline method to crowdworkers along with the corresponding dialogue context. Subsequently, these crowdworkers are requested to select which response is superior concerning the following five evaluation metrics.
\begin{description}
  \item[Consistency] The response is consistent with dialogue context.
  \item[Seeker's Knowledge] The response considers the seeker's level of knowledge.
  \item[Seeker's Interest] The response considers the seeker's level of interest.
  \item[Tailored Information] The response provides more information that the seeker does not know but is interested in.
  \item[Recommendation Success] The response is more likely to entice the seeker to watch the recommended movie.
\end{description}

Table~\ref{tab:human_evaluation_result} shows the win rates against the baseline method. Our proposed methods, CoT (sub) and CoT (obj), outperformed the baseline in all the metrics. Notably, the difference was statistically significant in Consistency, Seeker's Interest, and Recommendation Success for CoT (sub), and in Seeker's Knowledge, Seeker's Interest, Tailored Information for CoT (obj).

In addition, when correct labels were provided for the seeker's internal state estimation, there was a further improvement in the win rate. Notably, CoT (sub, gold) exhibited a higher win rate than CoT (obj, gold), indicating that considering the subjective (i.e., actual) seeker's internal state is effective in generating responses.

\subsection{Discussion}
In this section, we analyze the results of the seeker's internal state estimation, which is an intermediate task in our proposed framework. We consider the results divided into entity extraction and internal state classification.

\subsubsection{Entity Extraction}
We use precision and recall scores for exact matching as strict evaluation metrics and use the character-level F1 score as a lenient evaluation metric. To calculate the character-level F1 score, we first calculate the maximum character-level F1 score between each gold entity and the predicted entities. Then, we compute the average of these maximum values across all gold entities.

The precision and recall scores for the CoT (sub) model were observed to be 44.1 and 47.8 respectively, while the CoT (obj) model yielded scores of 42.7 and 46.3. These figures are relatively low, indicating a challenge in the model's ability to accurately estimate the precise spans of entities, particularly in terms of determining which modifiers should be included within the entity span. In contrast, the character-level F1 scores for the respective models exhibited higher values, achieving 76.2 and 76.1. This disparity in performance suggests that while the model encounters difficulties with precise entity span estimation, it is relatively adept at estimating approximate spans.

\subsubsection{Seeker's Internal State Classification}
\begin{table*}[t!]
  \centering
  \small
  \begin{tabular}{l|rrr|rrr}
  \toprule
  & \multicolumn{3}{c}{Knowledge} & \multicolumn{3}{c}{Interest} \\
  & \textit{High} & \textit{Neutral} & \textit{Low} & \textit{High} & \textit{Neutral} & \textit{Low} \\
  \midrule
  CoT (sub) & \textbf{74.2} & 9.9 & \textbf{49.5} & \textbf{84.7} & 23.1 & \textbf{26.9} \\
  Recommender & 70.4 & \textbf{14.4} & 46.4 & 76.3 & \textbf{27.6} & 25.5 \\
  \midrule
  CoT (obj) & \textbf{73.1} & 14.2 & \textbf{47.8} & \textbf{83.0} & 20.4 & \textbf{22.8} \\
  Recommender & 72.2 & \textbf{16.5} & 39.8 & 76.6 & \textbf{28.1} & 19.2 \\
  \bottomrule
  \end{tabular}
  \caption{Results of seeker's internal state classification.}
  \label{tab:internal_state_classification_result}
\end{table*}

We assess the classification performance of the seeker's internal state labels for successfully extracted entities using F1 score metric.

Table~\ref{tab:internal_state_classification_result} shows the results. In the context of knowledge and interest estimation, CoT (sub) and CoT (obj) demonstrated superior accuracy in predicting \textit{High} and \textit{Low} levels compared to human interlocutors (recommenders). However, for \textit{Neutral}, humans outperformed these models, indicating potential areas for further improvement. Additionally, when comparing CoT (sub) and CoT (obj), CoT (sub) generally achieved higher accuracy, suggesting the effectiveness of utilizing subjective labels.

Furthermore, knowledge and interest were estimated with relatively high accuracy for the \textit{High} category. Conversely, the \textit{Low} category exhibited lower accuracy, particularly regarding interest estimation. This lower performance is likely due to the imbalanced distribution of labels within the dataset. However, the primary focus of the study remains on the accurate identification of topics with \textit{High} interest in the context of recommendation dialogues rather than the identification of \textit{Low} interest topics. Consequently, this finding does not significantly detract from the overall utility of our proposed framework in recommendation scenarios.

\section{Conclusion}
We constructed RecMind, a recommendation dialogue dataset that features both subjective and objective annotations of the seeker's internal state at the entity level. Our dataset also has engaging dialogues with longer seeker's utterances, characterized by high scores in dialogue enjoyment and recommendation success. We also proposed a response generation framework that explicitly considers the seeker's internal state, applying Chain-of-Thought prompting to our task. The experimental results showed that our proposed method could generate responses that are more consistent and tailored to the seeker than the baseline method.

Our dataset has diverse and fine-grained annotations, which are useful for various tasks such as internal state estimation, external knowledge selection, and dialogue response generation. We hope our dataset will be useful for future research on recommendation dialogues.

\bibliography{custom}

\appendix
\section{Appendix}  \label{sec:appendix}

\subsection{Instruction for Workers} \label{sec:appendix_instruction}
We show the detailed instructions distributed to workers in this section.

\paragraph{Instruction for recommenders and seekers}
\begin{itemize}
  \item Do not participate in both roles in the same dialogue.
  \item Avoid dull and boring responses such as ``Yes'' and ``I see.''
  \item Avoid responses containing personal data.
  \item Avoid responses about this dialogue collection task itself.
  \item Do not use emoticons.
\end{itemize}

\paragraph{Instruction for recommenders only}
\begin{itemize}
  \item Select recommended movies from the movie search tool.
  \item May recommend movies that the seeker has already watched. In that case, however, try to recommend to make the seeker want to watch it again.
  \item Avoid too enthusiastically recommending movies you would like to recommend, ignoring the knowledge and interests of the seeker.
  \item Try to elicit sufficient information from the seeker and recommend movies you want that person to watch.
  \item Avoid short-sighted recommendations, such as ``Ask only the genre of the movie the seeker like (action, romance, etc.) and recommend one movie from that genre.''
\end{itemize}

\paragraph{Instruction for seekers only}
\begin{itemize}
  \item Actively ask questions about what you do not know or understand.
  \item Avoid requesting recommendations for recent movies (e.g., movies that are in theaters).
  \item Actively communicate what you know (or do not know) and what you are interested in (or not interested in) to the recommender.
\end{itemize}

\subsection{Dialogue Collection System Interface} \label{sec:appendix_system_interface}

\begin{figure*}[t!]
  \centering
  \includegraphics[width=\textwidth]{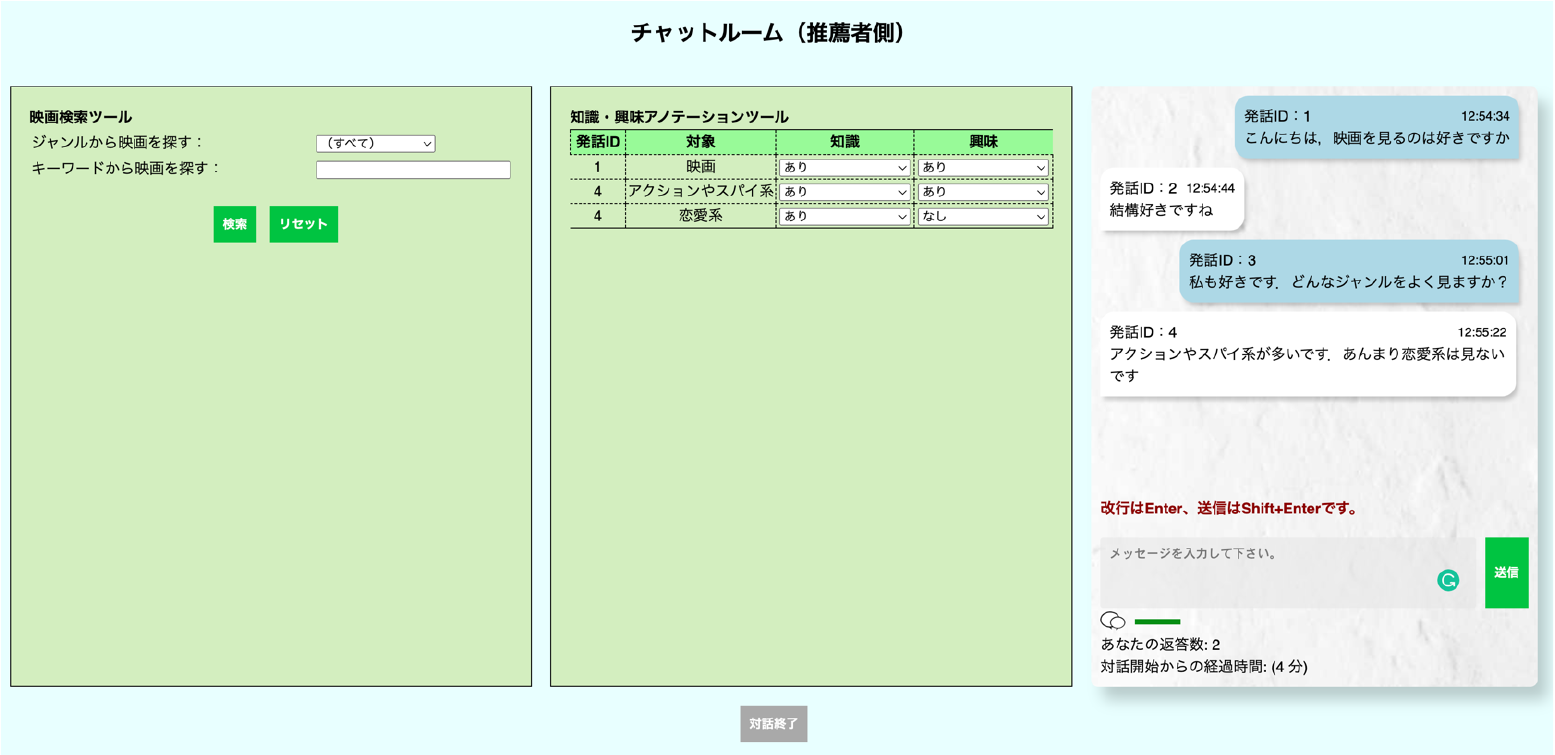}
  \caption{Screenshot of the recommender's chatroom}
  \label{fig:system_rec}
\end{figure*}

\begin{figure*}[t!]
  \centering
  \includegraphics[width=\textwidth]{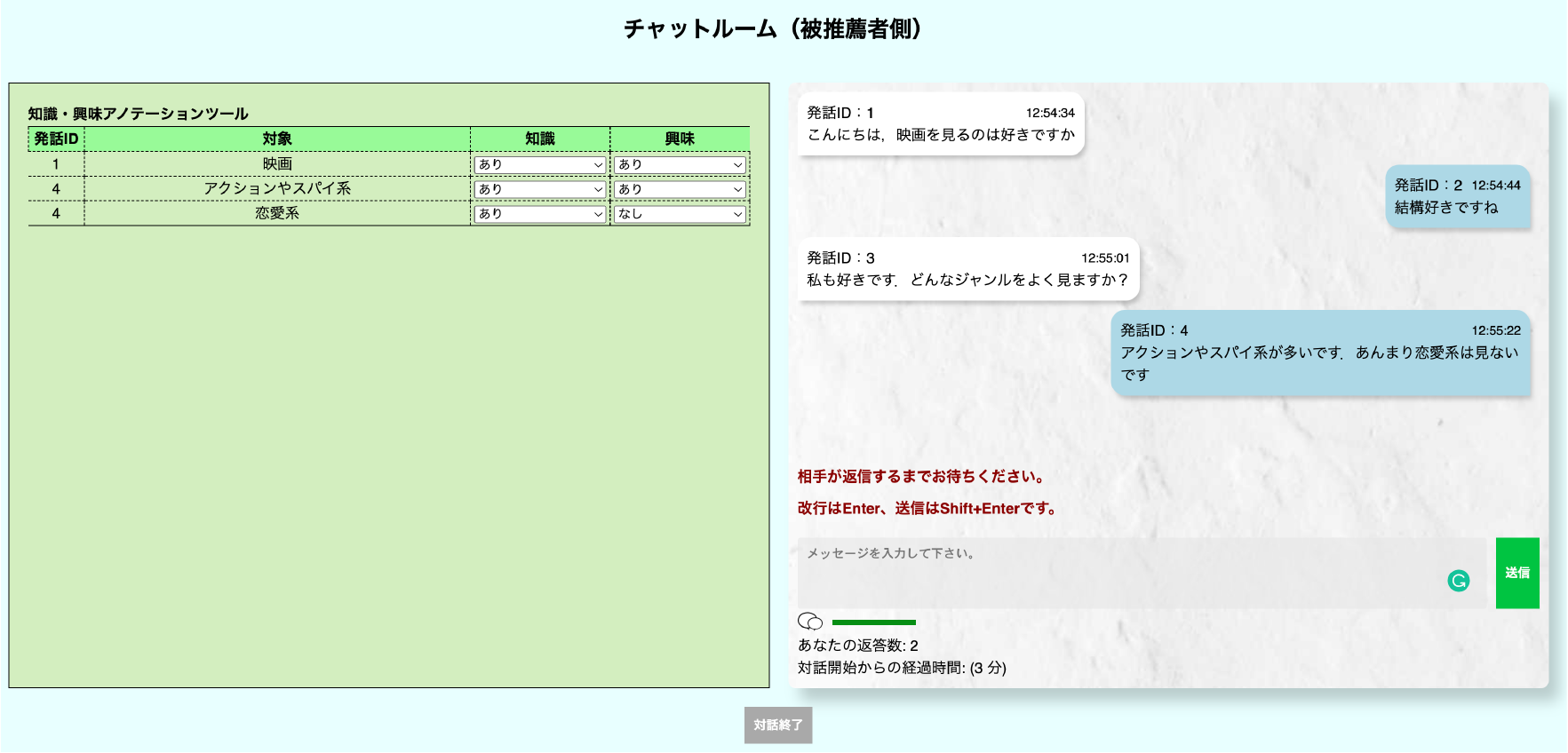}
  \caption{Screenshot of the seeker's chatroom}
  \label{fig:system_sek}
\end{figure*}

Figures~\ref{fig:system_rec} and \ref{fig:system_sek} show the screenshots of the dialogue collection system interface for the recommender and the seeker, respectively.

\end{document}